\documentclass{article} % For LaTeX2e
\usepackage{times}
\usepackage{multirow,xspace}
\usepackage{paralist} % Add all your packages here
\usepackage{url,enumerate}
\usepackage{color}
\usepackage{graphicx,epsfig,subfigure}
\usepackage{algorithm,algorithmic}
\usepackage{amsmath,amssymb}
\usepackage[numbers]{natbib}
\RequirePackage[colorlinks,citecolor=blue,urlcolor=blue]{hyperref}

\title{Infinite Tucker Decomposition: Nonparametric Bayesian Models for Multiway Data Analysis}

\newcommand{\infiTucker}{\textsl{InfTucker}\xspace}
\newcommand{\InfTucker}{\textsl{InfTucker}\xspace}
\newcommand{\itgp}{$\textsl{InfTucker}^{gp}$ \xspace}
\newcommand{\ittp}{$\textsl{InfTucker}^{tp}$ \xspace}
\newcommand{\TGP}{\mathcal{TGP}}
\newcommand{\TTP}{\mathcal{TTP}}
\newcommand{\TN}{\mathcal{TN}}
\newcommand{\TT}{\mathcal{TT}}
\newcommand{\PARAFAC}{CP\xspace}

\newcommand{\comment}[1]{}
\newcommand{\email}[1]{\href{mailto:#1}{#1}}

\input{./Definitions.tex}

\begin{document}

\author{
        Zenglin Xu \\
        Department of CS \\
        Purdue University \\
        \email{xu218@purdue.edu}{}
        \and
        Feng Yan \\
        Department of CS\\ Purdue University \\
        \email{yan12@purdue.edu}
        \and
        Yuan Qi \\
        Departments of CS and Statistics \\
        Purdue University \\
        \email{alanqi@cs.purdue.edu}
}

\maketitle

\begin{abstract}
% A fundamental method for analyzing multi-way data is tensor decomposition.
Tensor decomposition is a powerful computational tool for multiway data analysis.  Many popular tensor decomposition approaches---such as the Tucker decomposition and CANDECOMP/PARAFAC (CP)---amount to multi-linear factorization. They are insufficient to model (i) complex interactions between data entities, (ii) various data types (\eg missing data and binary data), and (iii) noisy observations and outliers. To address these issues, we propose tensor-variate latent nonparametric Bayesian models, coupled with efficient inference methods, for multiway data analysis. We name these models \infiTucker. Using these \infiTucker, we conduct Tucker decomposition in an infinite feature space. Unlike classical tensor decomposition models,  our new approaches handle both continuous and binary data in a probabilistic framework. Unlike previous Bayesian models on matrices and tensors, our models are based on latent Gaussian or $t$ processes with nonlinear covariance functions. To efficiently learn the \infiTucker from data, we develop a variational inference technique on tensors. Compared with classical implementation, the new technique reduces both time and space complexities by several orders of magnitude.
\comment{
This technique can be easily adopted in other contexts (\eg, multitask learning) where we also encounter tensor-variate Gaussian or $t$ processes.
}
Our experimental results on chemometrics and social network datasets demonstrate that our new models achieved significantly higher prediction accuracy than the most state-of-art tensor decomposition approaches.

\comment{

 But they are either designed
 for different tasks (\eg, multitask learning)----rather than multiway analysis---or use linear kernels to make the computation feasible in practice

a nonlinear probabilistic framework, called \infiTucker, which. It extends classical multi-linear factorization models to various data types by being capable of modeling general noises, \eg Gaussian noises and probit noises. To be robust to noises, \infiTucker involves tensor $t$ processes as  the underlying stochastic processes, which are essentially nonparametric priors on infinite tensor spaces.
%we propose \infiTucker---a nonparametric probabilistic framework for general $K$-mode Tucker decomposition in infinite feature spaces.
%Tensor $t$ processes, the underlying stochastic processes of \infiTucker, is essentially nonparametric priors on infinite tensors. Our \infiTucker framework also extends the data domain of classical multi-linear models by being capable of modeling general noises, \eg incomplete data and binary data.
As an important contribution, we propose a novel inference technique to overcome the computation barrier, which reduces both time and space complexities by several orders of magnitude. It thus leads to an efficient solution to the long standing high computation complexity problem for tensor $t$ processes and tensor Gaussian processes inferences.
%We propose a novel inference technique to overcome the computation barrier. Our algorithm provides a solution to the long standing high computation complexity problem for tensor $t$ processes and tensor Gaussian processes inferences, and both time and space complexities are reduced by several orders of magnitude.
Finally, empirical results suggest our model outperforms classical multi-linear tensor decomposition models on both regression and binary classification tasks on tensors.
} 
\end{abstract}

\section{Introduction}
%\comment{
%Modeling entity interactions is the central theme of data analysis in numerous scientific and engineering disciplines.
%
%Given increasingly more complex data that containing multiple aspects,  multiway data analysis has become an important research
%
%A factorization of data into low dimension spaces provides compact data representationes which allow us to concisely summarize data and examine the relations of the entities. Bilinear factor analysis, \eg Singular Value Decomposition (SVD) and Nonnegative Matrix Factorization (NMF), are successful for analyzing dyadic data. However,
%}

Many real-world datasets with multiple aspects can be described by tensors (i.e., multiway arrays). For example, email
correspondences can be represented by a tensor with four modes \verb"(sender, receiver, date, content)" and user
customer ratings by a tenor with four modes \verb"(user, item, rating, time)".
% Given the tensor-valued data,  how can we model them in a principled way and make accurate predictions?
% It is, however, a challenge task to conduct multiway analysis  due to complex interations,  missing data, outliers and thedata volume .
Given the tensor-valued data, traditional multiway factor models--- such as the Tucker decomposition \citep{Tucker66} and CANDECOMP/PARAFAC (CP) \citep{Harshman70parafac}---have been widely applied for various applications (\eg, network traffic analysis \citep{Zhang2009tensor}, computer vision \citep{Shashua05NTF} and social network analysis \citep{SunKDD06,Lin09Metafac, sun09multivis}, etc). These models, however, face serious challenges for modeling complex multiway interactions.  First the interactions between entities in each mode may be coupled together and highly nonlinear. The classical multi-linear models cannot capture these intricate relationships.
%Second, the tensor data are often sparse; they contain a lot of zero elements. The classical models often do not  model the data sparisity.  The classical models often do not  model the data sparisity.
Second, the data are often noisy, but the classical models are not designed to deal with noisy observations.
Third, the data may contain many missing values. We need to first impute the missing values before we can apply the classical multiway factor models. Forth, the data may not  be restricted to real values: they can be binary as in dynamic network data or have ordinal values for user-movie-ratings.  But the classical models simply treat them as continuous data---this treatment would lead to degenerated predictive performance.

To address these challenges we propose a nonparametric Bayesian multiway analysis model, \infiTucker. Based on latent Gaussian processes or $t$ processes, it conducts the Tucker decomposition in an infinite dimensional feature space.  It generalize the elegant work of \citet{Chu09ptucker} by capturing nonlinear interactions between different tensor modes.
% Compared with the tensor-variate Gaussian process approaches \citep{Hoff11multiway,Yu07SRM}, the $t$-process-based \infiTucker is more robust (not sensitive to data outliers).
Grounded in a probabilistic framework, it naturally handles noisy observations and missing data. Furthermore, it handles various data types---binary or continuous---by simply using suitable data likelihoods.
Although \infiTucker offers an elegant solution to multiway analysis, learning the model from data is computationally challenging. To overcome this challenge, we develop an efficient variational Bayesian approach that explores tensor structures to significantly reduce the computational cost. This efficient inference technique also enables the usage of nonlinear covariance functions for latent Gaussian and $t$ processes on datasets with reasonably large size.

Our experimental results on chemometrics and social network datasets demonstrate
that the \infiTucker achieves significantly higher prediction accuracy than state-of-the-art tensor decomposition approaches---including High Order Singular Value Decomposition (HOSVD) \citep{Lathauwer00HOSVD}, Weighted CP \citep{Acar11missing} and nonnegative tensor decomposition \citep{Shashua05NTF}.

%%\infiTucker defines a probabilistic \emph{infinite} Tucker decomposition model, whose marginalization on finite tensors is a \emph{tensor} $t$ distribution. The tensor processes induced by \infiTucker essentially defines a nonparametric prior on infinite tensors.

\comment{
In summary, the major contributions of this paper include:
\begin{compactenum}[i.]
\item We propose tensor-variate latent Gaussian and $t$ process models for multiway data analysis.
% Recently several nonparametric Bayesian models on matrices and tensors \citep{Chu09ptucker,Yu07SRM,Bonilla08multitask} have been developed, but they are either designed for different tasks (\eg, multitask learning)----rather than multiway analysis---or use a linear kernel to make the computation feasible in practice. By contrast,  based on latent $t$-processes, our model conducts robust nonlinear Tucker decomposition.

%\comment{
%% \infiTucker takes advantage of infinite feature mappings on low dimensional input spaces to model nonlinear interaction between entities, leading to nonlinear covariance functions between entities. Also, \infiTucker is applicable to various types of learning tasks and robust to outliers.
%    % due to its $t$ process nature \citep{Yu07tmultitask}.
%}

\item We develop an efficient variational inference on tensors by exploiting the Kronecker structure of the covariance.
%Specifically, for a $K$-mode tensor with  $n_k$ entities in each mode,
%(\alanc{Can we dont assume each mode has the same n???}),
%our algorithm has $O(\sum_{k=1}^K n_k^3+(\sum_{k=1}^K n_k)\prod_{k=1}^K n_k)$ time and $O(\sum_{k=1}^K n_k^2+\prod_{k=1}^K n_k)$ space complexities, instead of the $O(\prod_{k=1}^Kn_k^{3})$ time and $O(\prod_{k=1}^Kn_k^{2})$ space complexities of a na\"{i}ve inference technique.
The efficient inference technique enables us to use tensor-variate $t$-processes with nonlinear covariance functions.

%\alanc{1. I am not sure where you get n^{K+1}
%
%\alanc{2. The following content should be used earlier or in related work. "for tensor $t$ processes and for the closely related tensor Gaussian processes \citep{Bonilla08multitask}".
%}

\item Our experimental results on chemometrics and social network datasets demonstrate
that the \infiTucker achieves significantly higher prediction accuracy than several state-of-art tensor decomposition approaches including High Order Singular Value Decomposition (HOSVD) \citep{Lathauwer00HOSVD}, Weighted CP \citep{Acar11missing} and nonnegative tensor decomposition \citep{Shashua05NTF}.
\end{compactenum}
}

% Ref: MultiVis: Content-based Social Network Exploration Through Multiway Visual Analysis,

\section{Preliminary}
\label{sec:prelim}

{\bf Notations.} Throughout this paper, we denote scalars by lower case letters (\eg ~$a$),  vectors by bold lower case letters  (\eg ~$\mathbf{a}$), matrices by bold upper case letters (\eg ~$\mathbf{A}$), and tensors by calligraphic upper case letters (\eg ~$\mathcal{A}$). Calligraphic upper case letters are also used for probability distributions, $e.g.$, $\Ncal(\mub,\Sigmab)$.  We use $u_{ij}$ to represent the $(i,j)$ entry of a matrix $\Ub$, $y_{\vec{i}}$ to represent the $\vec{i}=(i_1,\ldots,i_K)$ entry of a tensor $\Ycal$. $\Ub\otimes\Vb$ denotes the Kronecker product of the two matrices there. We define the vectorization operation, denoted by $\vect(\Ycal)$, to stack the tensor entries into a $\prod_{k=1}^K n_k$ by 1 vector. The entry $\vec{i}=(i_1,\ldots,\i_K)$ of $\Ycal$ is mapped to the entry at position $j=i_K + \sum_{i=1}^{K-1} (i_k-1)\prod_{k+1}^K n_k$ of $\vect(\Ycal)$\footnote{\small Unlike the usual column-wise $\vect$-operation, our definition of $\vect()$ on matrices is row-wise, which avoids the use of transpose in many equations throughout this paper.}.
%When $K=2$, \ie matrix as 2-mode tensor, our definition of vectorization is the row-wise vectorization of matrix\footnote{The usual matrix $\vect$-operation is column-wise. Our definition avoids the use of transpose.}.
The mode-$k$ product of a tensor $\Wcal\in\RR^{r_1 \times\ldots\times r_K}$ with a matrix $\Ub\in\RR^{n\times r_k}$ is denoted as $\Wcal \times_k \Ub$ and it is of size $r_1\times\ldots\times r_{k-1}\times n \times r_{k+1}\times\ldots\times r_K$. The corresponding entry-wise definition is
\begin{align}
\label{eq:mode-k-product}
(\Wcal \times_k \Ub)_{i_1\ldots i_{k-1}ji_{k+1} \ldots i_K} = \sum_{j=1}^{r_k} w_{i_1 \ldots i_K} u_{ji_k}.
\end{align}
%
%Many of the above notations and definitions can be found in \citep{Kolda09TensorReview}.

{\bf Tensor decomposition:} There are two families of tensor decomposition, the Tucker family and the \PARAFAC family. The Tucker family extends bilinear factorization models to handle tensor datasets. For an \emph{observed} $K$-mode tensor $\Ycal\in\RR^{n_1\times\ldots\times n_K}$, the general form of Tucker decomposition is
\begin{equation}
\label{eq:tucker-decomp}
\Ycal = \Wcal \times_1 \Ub^{(1)} \times_2 \ldots \times_K \Ub^{(K)}
%\vect(\Wcal\times\Ucal) &= \Ub^{(1)}\otimes\Ub^{(2)}\otimes\ldots\otimes\Ub^{(K)} \cdot \vect(\Wcal)
\end{equation}
where $\Wcal\in\RR^{r_1 \times\ldots\times r_K}$ is the \emph{core} tensor, and $\Ub^{(k)}\in\RR^{n_k \times r_k}$ are $K$ latent factor matrices. As in \citep{Kolda09TensorReview}, We collectively denote the group of $K$ matrices as a Tucker tensor with a identity core $\Ucal=\left[\Ub^{(1)},\ldots,\Ub^{(K)}\right]$---this allows us to compactly represent the Tucker decomposition as $\Ycal=\Wcal\times\Ucal$. The vector form of \eqref{eq:tucker-decomp} is
\begin{equation}
\label{eq:tensor-vec-identity}
\vect(\Wcal\times\Ucal) = \Ub^{(1)}\otimes\Ub^{(2)}\otimes\ldots\otimes\Ub^{(K)} \cdot \vect(\Wcal)
\end{equation}
%
%The Tucker family models factorize the original observation tensor $\Ycal$ to $K$ components, and the interactions between components are described by the core tensor $\Wcal$.
%The solutions of Tucker decomposition can be obtained by solving the least-square problem $\min_{\Wcal,\Ub^{(k)}} \|\Ycal-\Wcal\times\Ucal\|^2$.
%
% \begin{align}
% \label{eq:tucker-optimize}
% \argmin_{\Wcal,\Ub^{(k)},k=1\ldots K} \|\Ycal-\Wcal\times\Ucal\|^2.
% \end{align}
%
The \PARAFAC family is a restricted form of the Tucker family. The entry-wise definition of \PARAFAC is $y_{i_1\ldots i_K} = \sum_{l=1}^r \lambda_l u_{i_1l}\ldots u_{i_Kl}$. The alternating least square (ALS) method has been used to solve both Tucker decomposition and \PARAFAC \citep{Kolda09TensorReview}.
%When $K=3$, the resulting Tucker decomposition is called Tucker-3 model. %\citet{Lathauwer00HOSVD} proposed High-Order Singular Value Decomposition (HOSVD) to preserve the orthogonality on $\Ub^{(k)}$ in the Tucker-3 model.

%A $K$-mode tensor $\Wcal$ is diagonal if and only if $r_1=\ldots=r_K=r$ and $w_{\vec{i}} \neq 0 \Longrightarrow i_1=\ldots=i_K$. The \PARAFAC (parallel factors) decomposition \citep{Harshman70parafac} is a constrained form of Tucker family decomposition with a diagonal core tensor. The \PARAFAC decomposition can be entry-wisely written as $y_{i_1\ldots i_K} = \sum_{l=1}^r \lambda_l u_{i_1l}\ldots u_{i_Kl}$.
%
% \begin{align}
% \label{eq:parafac-decomp}
% y_{i_1\ldots i_K} = \sum_{l=1}^m \lambda_l u_{i_1l}\ldots u_{i_Kl}
% \end{align}
%PARAFAC decomposition can be solved by ALS algorithm \citep{Kolda09TensorReview}. Unlike Tucker family decomposition, it is not possible to enforce orthogonality constraints on the component matrix $\Ub^{(k)}$ for PARAFAC due to the lack of degrees of freedom.

%%%%%%%%%%%%%%%%%%%%%%%%%%%%%%%%%%%%%%%%%%%%%%%%%%%%%%%%%%%%%
\comment{

{\bf Notations.} Throughout this paper, we denote a scalar by lower case letters, \eg, $x\in\RR$,  a vector by bold lower case letters, \eg, $\yb\in\RR^n$. Matrices are denoted by bold upper case letters, \eg, $\Ub$, and its transpose is denoted by $\Ub^\top$. A tensor (multiway array) is denoted by calligraphic upper case letters.
%, and the \emph{modes} of a tensor is the number of dimensions.
In this paper, we will be working with $K$-mode tensor $\Ycal\in\RR^{n_1 \times n_2\times\ldots\times n_K}$ with a dimension of $n_k$ at the $k$-th mode. As an exception, we use calligraphic upper case letters to denote probability distributions if they are followed by parameters, \eg, $\Ncal(\mub,\Sigmab)$ denotes the multivariate normal distribution with mean $\mub$ and covariance $\Sigmab$. Indices are denoted as subscripts, \eg, $u_{ij}$ is the $(i,j)$ entry of the matrix $\Ub$. We compactly denote a tensor index by $\vec{i}=(i_1,\ldots,i_K)$, and by $y_{\vec{i}}$ the corresponding entry in the tensor $\Ycal$. $\Ub\otimes\Vb$ denotes the Kronecker product of the two matrices $\Ub$ and $\Vb$. We define the vectorization operation for tensor, denoted by $\vect(\Ycal)$, to stack the entries in tensor $\Ycal$ into a $\prod_{k=1}^K n_k$ dimension vector. The entry $\vec{i}=(i_1,\ldots,\i_K)$ of $\Ycal$ is mapped to the entry at position $j=i_K + \sum_{i=1}^{K-1} (i_k-1)\prod_{k+1}^K n_k$ of $\vect(\Ycal)$\footnote{The usual matrix $\vect$-operation is column-wise. Our definition of $\vect()$ on matrices is row-wise, which avoids the use of transpose in many equations throughout this paper.}.
%When $K=2$, \ie matrix as 2-mode tensor, our definition of vectorization is the row-wise vectorization of matrix\footnote{The usual matrix $\vect$-operation is column-wise. Our definition avoids the use of transpose.}.
The mode-$k$ product of a tensor $\Wcal\in\RR^{r_1 \times\ldots\times r_K}$ with a matrix $\Ub\in\RR^{n\times r_k}$ is denoted as $\Wcal \times_k \Ub$ and it is of size $r_1\times\ldots\times r_{k-1}\times n \times r_{k+1}\times\ldots\times r_K$. The entry-wise definition is
$%\begin{align}
%\label{eq:mode-k-product}
(\Wcal \times_k \Ub)_{i_1\ldots i_{k-1}ji_{k+1} \ldots i_K} = \sum_{j=1}^{r_k} w_{i_1 \ldots i_K} u_{ji_k}.
$%\end{align}
%
%Many of the above notations and definitions can be found in \citep{Kolda09TensorReview}.

{\bf Tensor decomposition:} There are two families of tensor decomposition in the literatures, Tucker family and \PARAFAC family. The Tucker family decomposition extends bilinear factorization models to tensor datasets. For an \emph{observed} $K$-mode tensor $\Ycal\in\RR^{n_1\times\ldots\times n_K}$, the general form of Tucker decomposition is
\begin{equation}
\label{eq:tucker-decomp}
\Ycal = \Wcal \times_1 \Ub^{(1)} \times_2 \ldots \times_K \Ub^{(K)}
%\vect(\Wcal\times\Ucal) &= \Ub^{(1)}\otimes\Ub^{(2)}\otimes\ldots\otimes\Ub^{(K)} \cdot \vect(\Wcal)
\end{equation}
where $\Wcal\in\RR^{r_1 \times\ldots\times r_K}$ is the \emph{core} tensor, and $\Ub^{(k)}\in\RR^{n_k \times r_k}$ be $K$ latent factor matrices. We collectively denotes the group of $K$ matrices as a Tucker tensor with a identity core \citep{Kolda09TensorReview} $\Ucal=\left[\Ub^{(1)},\ldots,\Ub^{(K)}\right]$, and we compactly denote Tucker decomposition as $\Ycal=\Wcal\times\Ucal$. The vector form of \eqref{eq:tucker-decomp} is
\begin{equation}
\label{eq:tensor-vec-identity}
\vect(\Wcal\times\Ucal) = \Ub^{(1)}\otimes\Ub^{(2)}\otimes\ldots\otimes\Ub^{(K)} \cdot \vect(\Wcal)
\end{equation}
%
%The Tucker family decomposition model factorize the original observation tensor $\Ycal$ to $K$ components, and the interaction between components is described by the core tensor $\Wcal$. The solution of Tucker decomposition is obtained by solving the least-square problem $\min_{\Wcal,\Ub^{(k)}} \|\Ycal-\Wcal\times\Ucal\|^2$.
%
% \begin{align}
% \label{eq:tucker-optimize}
% \argmin_{\Wcal,\Ub^{(k)},k=1\ldots K} \|\Ycal-\Wcal\times\Ucal\|^2.
% \end{align}
%
The \PARAFAC family decomposition is a restricted form of the Tucker family decomposition. The entry-wise definition of \PARAFAC is $y_{i_1\ldots i_K} = \sum_{l=1}^r \lambda_l u_{i_1l}\ldots u_{i_Kl}$. The alternating least square (ALS) method can be employ to solve both Tucker decomposition and \PARAFAC \citep{Kolda09TensorReview}. %When $K=3$, the resulting Tucker family decomposition is called Tucker-3 model. \citet{Lathauwer00HOSVD} proposed High-Order Singular Value Decomposition (HOSVD) to preserve the orthogonality on $\Ub^{(k)}$ in the Tucker-3 model.

%A $K$-mode tensor $\Wcal$ is diagonal if and only if $r_1=\ldots=r_K=r$ and $w_{\vec{i}} \neq 0 \Longrightarrow i_1=\ldots=i_K$. The \PARAFAC (parallel factors) decomposition \citep{Harshman70parafac} is a constrained form of Tucker family decomposition with a diagonal core tensor. The \PARAFAC decomposition can be entry-wisely written as $y_{i_1\ldots i_K} = \sum_{l=1}^r \lambda_l u_{i_1l}\ldots u_{i_Kl}$.
%
% \begin{align}
% \label{eq:parafac-decomp}
% y_{i_1\ldots i_K} = \sum_{l=1}^m \lambda_l u_{i_1l}\ldots u_{i_Kl}
% \end{align}
%PARAFAC decomposition can be solved by ALS algorithm \citep{Kolda09TensorReview}. Unlike Tucker family decomposition, it is not possible to enforce orthogonality constraints on the component matrix $\Ub^{(k)}$ for PARAFAC due to the lack of degrees of freedom.

}% END of COmment 
\section{Infinite Tucker decomposition}
\label{sec:model}

In this section we present the infinite Tucker decomposition based on latent Gaussian processes and $t$ processes. The following discussion is primarily for latent Gaussian processes. The model derivation for  latent $t$ processes is similar to that of latent Gaussian processes.

We extend classical Tucker decomposition in three aspects: i) flexible noise models for both continuous and binary observations; ii) an infinite core tensor to model complex interactions; and iii) latent Gaussian process prior or latent $t$ process.
%, which makes the model robust to outliers.

More specifically, we assume the observed tensor $\Ycal$ is sampled from a latent real-valued tensor $\Mcal$ via a probabilistic noise model $p(\Ycal|\Mcal) = \prod_{\vec{i}} p(y_{\vec{i}}|m_{\vec{i}})$.
%
% \begin{align}
% \label{link-function}
% \Ycal\;|\;\Mcal \sim p(\Ycal|\Mcal) = \prod_{\vec{i}} p(y_{\vec{i}}|m_{\vec{i}})
% \end{align}
%

We conduct Tucker decomposition for $\Mcal$ with a core tensor $\Wcal$ of infinite size. To do so, we use a countably infinite feature mapping for the rows of the component matrix $\Ub^{(k)}\in\RR^{n_k\times r},~k=1,\ldots,K$. Let $\uu^{(k)}_i$ denotes the $i$-th row of $\Ub^{(k)}$, A feature mapping $\phi:\RR^r\rightarrow\RR^{\aleph_0}$ maps each $\uu^{(k)}_i$ to the infinite feature space $\phi(\uu^{(k)}_i)$, where $\aleph_0$ denotes the countable infinity. The inner product of the feature mapping is denoted as $\Sigmab^{(k)}_{ij} = \inner{\phi(\uu^{(k)}_i)}{\phi(\uu^{(k)}_j)}$. Let $\phi^{(r)}(\uu^{(k)}_i)=[\phi_1(\uu^{(k)}_i),\ldots,\phi_r(\uu^{(k)}_i)]$ denote the first $r$ coordinates of $\phi(\uu^{(k)}_i)$, $\Wcal\in\RR^{\aleph_0^K}$ denote an infinite $K$-mode core tensor, and $\Wcal^{(r)}=(w_{\vec{i}})_{i_k=1}^r\in\RR^{r^K}$ denote the first $r$ dimensions in every mode of $\Wcal$. The infinite Tucker decomposition ``$\Mcal=\Wcal\times\phi(\Ucal)$'' for the latent tensor $\Mcal$ can be formally defined as the limit of a series of finite Tucker decompositions.
\begin{align}
\label{eq:infinite-tucker-limit}
\Mcal = \lim_{r\rightarrow\infty} \Wcal^{(r)}\times_1\phi^{(r)}(\Ub^{(1)})\times_2\ldots\times_K\phi^{(r)}(\Ub^{(K)})
\end{align}
where $\phi^{r}(\Ub^{(k)})=[\phi^{(r)}(\uu^{(k)}_1)^\top,\ldots,\phi^{(r)}(\uu^{(k)}_{n_k})^\top]^\top$.

As shown in the next Section, we use a latent tensor-variate Gaussian process prior on $\Wcal$ and then marginalize it out to obtain a Gaussian process over $\Mcal$. Alternatively, we can also use a latent tensor-variate $t$ process prior on $\Wcal$ and obtain a $t$ process over $\Mcal$.
%---this makes the estimation of $\Mcal$ to the presence of strong noise and outliers.
% To formulate the model in a probabilistic way, we assign appropriate prior distribution on the core tensor $\Wcal$ (effectively on $\Wcal^{(r)}$). The model is nonparametric in the sense that the conditional model is infinite. But we integrating out $\Wcal$ to make the model tractable.

\subsection{Tensor-variate Gaussian processes}
Before formally defining the tensor-variate $t$ process, we denote the domain of the mode $k$ by $U_k$, the $K$ covariance functions by $\Sigma^{(k)}:U_k \times U_k\rightarrow\RR$, the covariance matrices by a Tucker tensor $\Scal^{-\frac{1}{2}}=[\rbr{\Sigmab^{(1)}}^{-\frac{1}{2}},\ldots,\rbr{\Sigmab^{(K)}}^{-\frac{1}{2}}]$ and $n=\prod_{k=1}^K n_k$. The norm of the a tensor $\|\Acal\|$ is defined as $\sqrt{\sum_{\vec{i}} a_{\vec{i}}^2}$. Then we define tensor-variate $t$ processes as follows.

\begin{definition}[Tensor-variate Gaussian Processes]
\label{def:tensor-process}
Given $K$ location sets $U_k$, $k=1,\ldots,K$, let $b:U_1\times\ldots\times U_K\rightarrow\RR$ be the mean function. $M=\{f(\uu^{(1)},\ldots,\uu^{(K)}) | \uu^{(k)} \in U_k\}$ is a set of random tensor variables where $f:U_1\times\ldots\times U_K\rightarrow\RR$ is a random function. For any finite sets $\{\uu^{(k)}_1,\ldots,\uu^{(k)}_{n_k}\}_{k=1}^K$, let $\Mcal=[f(\uu^{(1)}_{j_1},\ldots,\uu^{(K)}_{j_K})]_{\forall \vec{j}}\in \RR^{n_1\times\ldots\times n_K} $, where $j_k=1,\ldots,n_k$, be a random tensor and $\Bcal=[b(\uu^{(1)}_{j_1},\ldots,\uu^{(K)}_{j_K})]_{\forall \vec{j}} \in \RR^{n_1\times\ldots\times n_K}$ be the mean tensor.

We say $\Mcal\sim \TGP(\Mcal|\Bcal,\{\Sigmab^{(k)}\}_{k=1}^K)$ follows a tensor-variate Gaussian process, if $\Mcal$ follows a tensor-variate normal distribution:
\begin{align}
\label{eq:gaussian-process-pdf}\small
\TN&(\Mcal|\Bcal,\{\Sigmab^{(k)}\}_{k=1}^K) = (2\pi)^{-\frac{n}{2}} \prod_{k=1}^K  |\Sigmab^{(k)}|^{-\frac{n}{2n_k}}\nonumber \\
&\exp\cbr{-\frac{1}{2} \|(\Mcal-\Bcal) \times\Scal^{-\frac{1}{2}}\|^2}.
\end{align}
\end{definition}

%
%\begin{definition}
%We say $\Mcal\sim \TGP(\Mcal|\Bcal,\{\Sigmab^{(k)}\}_{k=1}^K)$ follows a tensor-variate Gaussian process, if $\Mcal$ follows a tensor-variate normal distribution:
%\begin{align}
%\label{eq:gaussian-process-pdf}\small
%\TN&(\Mcal|\Bcal,\{\Sigmab^{(k)}\}_{k=1}^K) = (2\pi)^{-\frac{n}{2}} \prod_{k=1}^K  |\Sigmab^{(k)}|^{-\frac{n}{2n_k}}\nonumber \\
%&\exp\cbr{-\frac{1}{2} \|(\Mcal-\Bcal) \times\Scal^{-\frac{1}{2}}\|^2}.
%\end{align}
%\end{definition}
%
%We can also prove a similar convergence result for tensor-variate Gaussian distribution.
%%
%\begin{theorem}
%\label{thm:tensor-process-convergence2}
%% Let $U_k \subset \RR^r$, and $\Sigma^{(k)}_r(\uu^{(k)}_i,\uu^{(k)}_j)=\inner{\phi^{(r)}(\uu^{(k)}_i)}{\phi^{(r)}(\uu^{(k)}_j)}$ be a series of covariance functions. Define a multi-linear function by
%% \begin{align*}
%% t^{(r)}(\uu^{(1)},\ldots,\uu^{(K)})=\Wcal^{(r)}\times_1\phi^{(r)}(\uu^{(1)})\ldots\times_K\phi^{(r)}(\uu^{(K)}),
%% \end{align*}
%% where $\uu^{(k)}\in U_k$.
%If $\Wcal^{(r)} \sim \TN(\0,\{\Ib_r\}_{k=1}^K)$, then $t^{(r)}(\uu^{(1)},\ldots,\uu^{(K)})$ follows a tensor-variate Gaussian distribution $\TN(\0,\{\Sigma^{(k)}_r\}_{k=1}^K)$, and it converges to tensor-variate Gaussian processes $\TGP(0,\{\Sigma^{(k)}\}_{k=1}^K)$ in distribution as ${r\rightarrow\infty}$.
%\end{theorem}

In this paper, we set the mean function to be zero, \ie $\Bcal=0$.
%It is straight-forward to verify that our definition of tensor processes satisfies the consistency condition, \ie given finite projection onto a set of finite locations, its marginal distribution on a subset of the locations is the same as direct projection.
Let $\Ncal(\nu,\mub,\Sigmab)$ denotes a normal distribution with mean $\mub$ and covariance $\Sigmab$. If the latent tensor $\Mcal$ is drawn from a tensor-variate Gaussian process, then $\vect(\Mcal) \sim \Ncal(\0,\Psib)$, where $\Psib=\Sigmab^{(1)}\otimes\ldots\otimes\Sigmab^{(K)}$. We choose the prior on the truncated core tensor $\Wcal^{(r)}$ to be $\TN(\0,\{\Ib_r\}_{k=1}^K)$, where $\Ib_r$ denotes the identity matrix. The next theorem proves that the limit defined in \eqref{eq:infinite-tucker-limit} is the corresponding tensor process.
\begin{theorem}
\label{thm:tensor-process-convergence}
Let $U_k \subset \RR^r$, and $\Sigma^{(k)}_r(\uu^{(k)}_i,\uu^{(k)}_j)=\inner{\phi^{(r)}(\uu^{(k)}_i)}{\phi^{(r)}(\uu^{(k)}_j)}$ be a series of covariance functions. Define a multi-linear function by
\begin{align*}
g^{(r)}(\uu^{(1)},\ldots,\uu^{(K)})=\Wcal^{(r)}\times_1\phi^{(r)}(\uu^{(1)})\ldots\times_K\phi^{(r)}(\uu^{(K)}),
\end{align*}
where $\uu^{(k)}\in U_k$.
%where $\Wcal^{(r)}\in\RR^{m^K}$ is the core tensor.
%\begin{enumerate}[i.]
%\item
%If $\Wcal^{(r)} \sim \TN(\0,\{\Ib_m\}_{k=1}^K)$, then $t^{(r)}(\uu^{(1)},\ldots,\uu^{(K)})$ follows a tensor Gaussian distribution $\TGP(\0,\{\Sigma^{(k)}_m\}_{k=1}^K)$, and it converges to $\TGP(\0,\{\Sigma^{(k)}\}_{k=1}^K)$ in distribution as ${m\rightarrow\infty}$.
%
%\item
If $\Wcal^{(r)} \sim \TN(\nu,\0,\{\Ib_r\}_{k=1}^K)$, then $g^{(r)}(\uu^{(1)},\ldots,\uu^{(K)})$ follows a tensor-variate Gaussian distribution $\TN(\0,\{\Sigma^{(k)}_r\}_{k=1}^K)$, and it converges to $\TGP(\0,\{\Sigma^{(k)}\}_{k=1}^K)$ in distribution as ${r\rightarrow\infty}$.
%\end{enumerate}
\end{theorem}
The proof of Theorem \ref{thm:tensor-process-convergence2} can be found in Appendix \ref{subsec:proof}.

%The proof of Theorem \ref{thm:tensor-process-convergence2} follows exactly the same path as that of the convergence result for tensor-variate $t$ processes.

Finally, to encourage sparsity in estimated $\uu^{(k)}_i$---for easy model interpretation---we use Laplace prior $\uu^{(k)}_i\sim\Lcal(\lambda)\propto\exp(-\lambda\|\uu^{(k)}_i\|_1)$.

\comment{
It is customary to place priors on $\uu^{(k)}_i$ to specify a fully generative model. The Gaussian prior, $\uu^{(k)}_i\sim \Ncal(\0,\lambda\Ib)$, constrains the complexity of the model, which is favorable for optimization. The Laplace prior, $\uu^{(k)}_i\sim\Lcal(\lambda)\propto\exp(-\lambda\|\uu^{(k)}_i\|_1)$, induces sparse solutions. We use Laplace prior in this paper.
}

\subsection{Tensor-variate $t$ processes}

Because of the strong relation between $t$-distributions and Gaussian distributions---$t$ distributions can be regarded as mixtures of Gaussian distributions weighted by Gamma distributions, we can easily define tensor-variate $t$ processes:

\begin{definition}[Tensor-variate $t$ Processes]
%\begin{definition}[Tensor-variate $t$ Distributions and Tensor-variate $t$ Processes]
\label{def:tensor-process}
Let $\Gamma(x)$ be the Gamma function. The set $M$ follows a tensor-variate $t$ process $\TTP(\nu, b,\{\Sigma^{(k)}\}_{k=1}^K)$ with degree of freedom $\nu>2$, if $\Mcal$ follows \emph{tensor $t$} distribution with the following density
%
%\begin{align}
\begin{eqnarray*}
\label{eq:t-process-pdf}
&\TT&(\Mcal|\nu,\Bcal,\{\Sigmab^{(k)}\}_{k=1}^K) = \frac{\Gamma(\frac{n+\nu}{2})\prod_{k=1}^K  |\Sigmab^{(k)}|^{-\frac{n}{2n_k}}}{\Gamma(\frac{\nu}{2})(\nu\pi)^{\frac{n}{2}}}\\
&&\rbr{1 + \frac{1}{\nu} \|(\Mcal-\Bcal) \times\Scal^{-\frac{1}{2}}\|^2}^{-\frac{1}{2}(n+\nu)}
\end{eqnarray*}
%\end{align}
%\end{enumerate}
%
\end{definition}

We can also prove a similar convergence result for tensor-variate Gaussian distribution.
\begin{theorem}
\label{thm:tensor-process-convergence2}
% Let $U_k \subset \RR^r$, and $\Sigma^{(k)}_r(\uu^{(k)}_i,\uu^{(k)}_j)=\inner{\phi^{(r)}(\uu^{(k)}_i)}{\phi^{(r)}(\uu^{(k)}_j)}$ be a series of covariance functions. Define a multi-linear function by
% \begin{align*}
% t^{(r)}(\uu^{(1)},\ldots,\uu^{(K)})=\Wcal^{(r)}\times_1\phi^{(r)}(\uu^{(1)})\ldots\times_K\phi^{(r)}(\uu^{(K)}),
% \end{align*}
% where $\uu^{(k)}\in U_k$.
If $\Wcal^{(r)} \sim \TT(\nu,\0,\{\Ib_r\}_{k=1}^K)$, then $g^{(r)}(\uu^{(1)},\ldots,\uu^{(K)})$ follows a tensor-variate $t$ distribution $\TT(\nu,\0,\{\Sigma^{(k)}_r\}_{k=1}^K)$, and it converges to tensor-variate $t$ processes $\TTP(\nu,0,\{\Sigma^{(k)}\}_{k=1}^K)$ in distribution as ${r\rightarrow\infty}$.
\end{theorem}

The proof of Theorem \ref{thm:tensor-process-convergence} follows exactly the same path as that of the convergence result for tensor-variate Gaussian processes.

The above theorem shows that probabilistic infinite Tucker decomposition of $\Mcal$ can be realized by modeling $\Mcal$ as a draw from a tensor-variate $t$ process on the location vectors induced from the unknown component matrices $\Ub^{(k)}$. Our definition of tensor-variate $t$ processes generalizes matrix-variate $t$ process defined in \citep{Zhang10multitask}. Theorem \ref{thm:tensor-process-convergence} also suggests a constructive definition of tensor-variate processes for general covariance functions.

\subsection{Noise models}

We use a noise model  $p(\Ycal|\Mcal)$ to link the infinite Tucker decomposition and the tensor observation $\Ycal$.

{\bf Probit model:} In this case, each entry of the observation is binary; that is, $y_{\vec{i}}\in\{0,1\}$. A probit function $p(y_{\vec{i}}|m_{\vec{i}})=\Phi(m_{\vec{i}})^{y_{\vec{i}}}(1-\Phi(m_{\vec{i}}))^{1-y_{\vec{i}}}$ models the binary observation. Note that $\Phi(\cdot)$ is the standard normal cumulative distribution function.

{\bf Gaussian model:} We use a Gaussian likelihood $p(y_{\vec{i}}|m_{\vec{i}}) = \Ncal(y_{\vec{i}}|m_{\vec{i}},\sigma^2)$
to model the real-valued observation $y_{\vec{i}}$.

{\bf Missing values:} We allow missing values in the observation. Let $\OO$ denote the indices of the observed entries in $\Ycal$. Then we have  $p(\Ycal_\OO|\Mcal_\OO)$ as the likelihood.

Other noise models include modified probit models for ordinal regression and multi-class classification \citep{Albert_Chib_1993Probit}, null category noise models for semi-supervised classification \citep{Lawrence05semi}. In this paper we focus on probit and Gaussian models. 
\section{Algorithm}
\label{sec:inference}

Given the observed tensor $\Ycal$, we aim to estimate the component matrices $\Ub^{(k)}$  by maximizing the marginal likelihood %$p(\{\Ub^{(k)}\}_{k=1}^K|\Ycal) \propto
$p(\Ycal|\{\Ub^{(k)}\}_{k=1}^K)p(\{\Ub^{(k)}\}_{k=1}^K).$
 Integrating out $\Mcal$ in the above equation is intractable however. Therefore, we resort to approximate inference; more specifically, we develop a variational expectation maximization (EM) algorithm. In the following paragraphs, we first present the inference and prediction algorithms for both of the noise models, and then describe an efficient algebraic approach to significantly reduce the computation complexity. Due to space limitation, we only describe the algorithm for tensor-variate $t$-distribution. The algorithm for tensor-variate Gaussian distribution can be derived similarly. %Due to the relation \eqref{eq:t-distribution-fac} between normal distribution and $t$ distribution, we only present the algorithms for $\Mcal$ drawn from tensor $t$ processes. The inference algorithms for tensor Gaussian processes follow a similar line.

\subsection{Inference}

{\bf Probit noise:}  We follow the data augmentation scheme by \citet{Albert_Chib_1993Probit} to decompose the probit model into $p(y_{\vec{i}}|m_{\vec{i}})=\int p(y_{\vec{i}}|z_{\vec{i}})p(z_{\vec{i}}|m_{\vec{i}})dz_{\vec{i}}$ . Let $\delta(\cdot)$ be the indicator function, we have
\begin{align*}
p(y_{\vec{i}}|z_{\vec{i}}) &= \delta(y_{\vec{i}}=1)\delta(z_{\vec{i}}>0)+\delta(y_{\vec{i}}=0)\delta(z_{\vec{i}}\leq 0),\\
p(z_{\vec{i}}|m_{\vec{i}}) &= \Ncal(z_{\vec{i}}|m_{\vec{i}},1)
\end{align*}
It is well known that a $t$ distribution can be factorized into a normal distribution convolved with a Gamma distribution, such that
\begin{align}
\label{eq:t-tensor-fac}
\TT&(\Mcal|\nu,\0,\{\Sigmab^{(k)}\}_{k=1}^K) = \int  \textrm{Gam}(\eta|\nu/2,\nu/2) \cdot\nonumber\\
&\TN(\Mcal|\0,\{\eta^{-1/K}\Sigmab^{(k)}\}_{k=1}^K)d\eta, %\\
%\TN(\Mcal|\Bcal,\{\Sigmab^{(k)}\}_{k=1}^K) & = & (2\pi)^{-\frac{n}{2}} \prod_{k=1}^K  |\Sigmab^{(k)}|^{-\frac{n}{2n_k}}\exp\left(-\frac{1}{2} \|(\Mcal-\Bcal) \times\Scal^{-\frac{1}{2}}\|^2\right),
\end{align}
where $\TN$ denotes the tensor-variate normal distribution. The joint probability likelihood with data augmentation is
\begin{align}
\label{eq:joint-likelihood}
p(\Ycal,\Zcal,\Mcal,\eta,\Ucal)=p(\Ycal|\Zcal)p(\Zcal|\Mcal)p(\Mcal|\eta,\Ucal)p(\eta)p(\Ucal).
\end{align}
where $p(\Mcal|\eta,\Ucal)$ and $p(\eta)$ is the tensor-variate normal distribution and the Gamma distribution in \eqref{eq:t-tensor-fac}. $p(\Ucal)$ is the Laplace prior.

Our variational EM algorithm consists of a variational E-step and a gradient-based M-step. In the E-step, we approximate the posterior distribution $p(\Zcal,\Mcal,\eta|\Ycal,\Ucal)$ by a fully factorized distribution $q(\Zcal,\Mcal,\eta)=q(\Zcal)q(\Mcal)q(\eta)$. Variational inference minimizes the Kullback-Leibler (KL) divergence between the approximate posterior and the true posterior.
\begin{align}
\label{eq:KL-divergence}
\min_{q}\textrm{KL}\rbr{q(\Zcal)q(\Mcal)q(\eta)\|p(\Zcal,\Mcal,\eta|\Ycal,\Ucal)}.
\end{align}
The variational approach optimizes one approximate distribution, \eg, $q(\Zcal)$, in \eqref{eq:KL-divergence} at a time, while having all the other approximate distributions fixed \citep{Bishop07PRML}. We loop over $q(\Zcal)$, $q(\Mcal)$ and $q(\eta)$ to iteratively optimize the KL divergence until convergence.

Given $q(\Mcal)$ and $q(\eta)$, the $q(z_{\vec{i}})$ is a truncated normal distribution
\begin{align}
\label{eq:Z-optimized}
&q(z_{\vec{i}}) \propto \Ncal(\expec{q}{m_{\vec{i}}},1)\delta(z_{\vec{i}}>1),\\
&\expec{q}{z_{\vec{i}}}=\expec{q}{m_{\vec{i}}} + \frac{(2y_{\vec{i}}-1)\Ncal(\expec{q}{m_{\vec{i}}}|0,1)}{\Phi((2y_{\vec{i}}-1)\expec{q}{m_{\vec{i}}})}.
\end{align}
Given $q(\Zcal)$ and $q(\eta)$, it is more convenient to write the optimized approximate distribution for $\Mcal$ in its vectorized form. Let $\Sigmab_p = \Sigmab^{(1)}\otimes\ldots\otimes\Sigmab^{(K)}$, we have
\begin{align}
q(\vect(\Mcal)) &= \Ncal(\vect(\Mcal)|\mub, \Upsilonb),\\
\mub &= \vect(\expec{q}{\Mcal}) = \Upsilonb~\vect(\expec{q}{\Zcal}) \label{eq:M-optimized} \\
\Upsilonb &= \expec{q}{\eta}^{-1} \Sigmab_p \rbr{\Ib + \expec{q}{\eta}^{-1}\Sigmab_p}^{-1}.\label{eq:Upsilonb_def}
\end{align}
The optimized $q(\eta)$ is also a Gamma distribution: %, while keeping the other two approximate distribution fixed.
\begin{align*}
%\label{eq:eta-optimized}
&q(\eta) = \textrm{Gam}(\eta|\beta_1,\beta_2),~~\expec{q}{\eta} = \frac{\beta_1}{\beta_2},~~
\beta_1 = \frac{\nu + n}{2},\\
&\beta_2 = \frac{\nu + \mub^\top \Sigmab_p^{-1} \mub +\tr(\Sigmab_p^{-1}\Upsilonb)}{2}.
\end{align*}

Based on the variational approximate distribution obtained in the E-step, we maximize the expected log likelihood over $\Ucal = \sbr{\Ub^{(1)},\ldots,\Ub^{(K)}}$ in the M-step.
\begin{align}
\label{eq:M-step-log-likelihood}
\max_{\Ucal} \expec{q}{\log p(\Ycal,\Zcal,\Mcal,\eta|\Ucal)p(\Ucal)}.
\end{align}
After eliminating constant terms, we need to solve the following optimization problem
\begin{align}
\min_{\Ucal} f(\Ucal)& = \sum_{k=1}^K \frac{n}{n_k}\log|\Sigmab^{(k)}| +  \tau \|\expec{q}{\Mcal}\times\Scal^{-1/2}\|^2 \nonumber\\
&+ \tau \tr\rbr{\Sigmab_p^{-1}\Upsilonb} + \lambda\sum_{k=1}^K \|\Ub_k\|_1,
\label{eq:optimization-M-step}
\end{align}
where $\tau=\expec{q}{\eta}$.
% \mub^\top\Sigmab_p^{-1}\mub
%
In the above equation \eqref{eq:optimization-M-step}, $\Sigmab^{(k)}=\Sigma^{(k)}(\Ub^{(k)},\Ub^{(k)})$ is considered as a function of $\Ub_k$, and $\Scal^{-1/2}$ is a function of $\Ucal$. $\Upsilonb$ and $\tau$ are the statistics computed in the E-step, and they have fixed values. The gradient of $f(\Ucal)$ \wrt to a scalar $u^{(k)}_{ij}$ can be found in Appendix \ref{subsec:grad}.
%Omitting the last $\ell_1$ term in \eqref{eq:optimization-M-step}, the gradient of $f(\Ucal)$ \wrt to a scalar $u^{(k)}_{ij}$ is the following.
%%
%\begin{align}
%\frac{\partial{f}}{\partial{u^{(k)}_{ij}}} = \;&\frac{n}{n_k} \tr\rbr{(\Sigmab^{(k)})^{-1} \frac{\partial\Sigmab^{(k)}}{\partial{u^{(k)}_{ij}}}} \nonumber\\
%&+ \tau \mub^\top\Deltab^{(k)}\mub + \tau \tr\rbr{\Deltab^{(k)}\Upsilonb} \\
%\Deltab^{(k)} = \;&(\Sigmab^{(1)})^{-1}\otimes\ldots\otimes(\Sigmab^{(k-1)})^{-1}\nonumber\\&\otimes (\Sigmab^{(k)})^{-1} \frac{\partial\Sigmab^{(k)}}{\partial{u^{(k)}_{ij}}} (\Sigmab^{(k)})^{-1}  \nonumber\\ &\otimes(\Sigmab^{(k+1)})^{-1}\otimes\ldots\otimes(\Sigmab^{(K)})^{-1} \nonumber
%\end{align}
%
With an $\ell_1$ penalty on  $f(\Ucal)$, we choose a projected scaled subgradient L-BFGS algorithm for optimization---due to its excellent performance \citep{Schmidt10}.
%Finally, We use $\expec{q}{\Mcal}$ as predictions to calculate the area-under-curve value.

{\bf Gaussian noise:} The inference for the regression case follows the same format as the binary classification case. The only changes are: 1) replacing $\expec{q}{\Zcal}$ by $\Ycal$ and skipping updating $q(\Zcal)$. 2) The variational EM algorithm are only applied to the observed entries. %If $\Mcal$ is drawn from tensor Gaussian process, we can also integrate out $\Mcal$ and maximize the log likelihood directly.

\subsection{Prediction}

{\bf Probit noise:} Given a missing value index $\vec{i}=(i_1,\ldots,i_K)$, the predictive distribution is
\begin{align}
&p(y_{\vec{i}}=1|\Ycal) \approx \nonumber\\ &\int p(y_{\vec{i}}=1|m_{\vec{i}})p(m_{\vec{i}}|\Mcal,\eta)q(\Mcal)q(\eta)m_{\vec{i}}d\Mcal d\eta
\label{eq:prediction}
\end{align}
The above integral is intractable, so we replace $\eta$ integral $q(\eta)d\eta$ by the mode of its approximate posterior distribution $\tau^*=(\beta_1-1)/\beta_2$, thus the predictive distribution is approximated by
\begin{align}
&\int p(y_{\vec{i}}=1|z_{\vec{i}}) p(z_{\vec{i}}|m_{\vec{i}})p(m_{\vec{i}}|\Mcal,\tau^*)q(\Mcal)dz_{\vec{i}}dm_{\vec{i}}d\Mcal \nonumber\\
&= \; \int \delta(z_{\vec{i}}>0) \Ncal(z_{\vec{i}}|\mu_{\vec{i}}(1), \nu_{\vec{i}}^2(1) )dz_{\vec{i}} \nonumber \\
&= \Phi(\frac{\mu_{\vec{i}}(1)}{\nu_{\vec{i}}(1)}) \label{eq:predictive-dist-probit}
\end{align}
where  %For tensor Gaussian process, there is no $\tau^*$ term on the covariance (equivalently $\tau^*=1$), and the integral \eqref{eq:predictive-dist} is the exact predictive distribution.
\begin{align*}
&k(\vec{i},\vec{j})=\prod_{k=1}^K \Sigma^{(k)}(\uu^{(k)}_{i_k},\uu^{(k)}_{j_k}), \;\kb=\sbr{k(\vec{i},\vec{j})}^\top_{\vec{j}\in\OO}\\
&\mu_{\vec{i}}(\rho) = \kb^\top(\Sigmab_p+\rho^2\tau^*\Ib)^{-1} \vect(\Ycal) \\
&\nu_{\vec{i}}^2(\rho) = 1 + \frac{1}{\tau^*}[k(\vec{i},\vec{i})- \kb^\top(\Sigmab_p+\rho^2\tau^*\Ib)^{-1}\kb]
\end{align*}

{\bf Gaussian noise:} The predictive distribution for the regression case is the following integral
\begin{align}
p(y_{\vec{i}}|\Ycal_\OO) \approx &\int p(y_{\vec{i}}|m_{\vec{i}})p(m_{\vec{i}}|\Mcal,\eta)q(\Mcal)q(\eta)m_{\vec{i}}d\Mcal d\eta \nonumber \\
\approx & \int p(y_{\vec{i}}=1|m_{\vec{i}})p(m_{\vec{i}}|\Mcal,\tau^*)q(\Mcal)dz_{\vec{i}}dm_{\vec{i}}d\Mcal \nonumber\\
= & \Ncal(z_{\vec{i}}|\mu_{\vec{i}}(\sigma), \nu_{\vec{i}}^2(\sigma) ).
\end{align}

\subsection{Efficient Computation}\label{sec:inf:comp}

A na\"{i}ve implementation of the above algorithm requires prohibitive $O(\prod_{k=1}^K n_k^3)$ time complexity and $O(\prod_{k=1}^K n_k^2)$ space complexity for each EM iteration. The key computation bottlenecks are the operations involving $\Upsilonb$ defined in equation (\ref{eq:Upsilonb_def}). To avoid this high complexity, we can make use of the Kronecker product structure. We assume $\expec{q}{\eta}=1$ to simplify the computation, it is easy to adapt our computation strategies to $\expec{q}{\eta} \neq 1$. Let $\Sigmab^{(k)}=\Vb^{(k)}\Lambdab^{(k)}\Vb^{(k)\top}$ be the singular value decomposition of the covariance matrix $\Sigmab^{(k)}$, $\Vb^{(k)}$ is an orthogonal matrix and $\Lambdab^{(k)}$ is a diagonal matrix. $\Upsilonb$ can be represented as
%
%\begin{align*}
%\Upsilonb = \;&\Vb^{(1)}\Lambdab^{(1)}\Vb^{(1)\top}\otimes\ldots\otimes\Vb^{(K)}\Lambdab^{(K)}\Vb^{(K)\top} (\Vb^{(1)}\Vb^{(1)\top}\otimes\ldots\otimes\Vb^{(K)}\Vb^{(K)\top} + \\ &\Vb^{(1)}\Lambdab^{(1)}\Vb^{(1)\top}\otimes\ldots\otimes\Vb^{(K)}\Lambdab^{(K)}\Vb^{(K)\top})^{-1} \\
%= \; & \Vb^{(1)}\Lambdab^{(1)}(\Ib+\Lambdab^{(1)})^{-1}\Vb^{(1)\top}\otimes
%\ldots\otimes\Vb^{(K)}\Lambdab^{(K)}(\Ib+\Lambdab^{(K)})^{-1}\Vb^{(K)\top}.
%\end{align*}

\begin{align*}
\Upsilonb = & \Vb^{(1)}\Lambdab^{(1)}(\Ib+\Lambdab^{(1)})^{-1}\Vb^{(1)\top}\otimes
\ldots\otimes\Vb^{(K)} \\
& \Lambdab^{(K)}(\Ib+\Lambdab^{(K)})^{-1}\Vb^{(K)\top}.
\end{align*}
Let $\Vb=\Vb^{(1)}\otimes\ldots\otimes\Vb^{(K)}$, $\Lambdab=\Lambdab^{(1)}(\Ib+\Lambdab^{(1)})^{-1}\otimes\ldots\otimes\Lambdab^{(K)}(\Ib+\Lambdab^{(K)})^{-1}$. It is obvious that $\Vb$ is an orthogonal matrix and $\Lambdab$ is a diagonal matrix. The above relation implies that we can actually compute the singular value decomposition of $\Upsilonb=\Vb\Lambdab\Vb^\top$ from covariance matrices $\Sigmab^{(k)}$.

In order to efficiently compute $\tr(\Sigmab_p^{-1}\Upsilonb)$ appearing in equation (\ref{eq:optimization-M-step}), we use the following relations
\begin{align}
\tr&(\Sigmab_p^{-1}\Upsilonb)=\tr(\Sigmab_p^{-1}\Vb^\top\Lambdab\Vb)=\tr(\Lambdab\Vb\Sigmab_p^{-1}\Vb^\top) \nonumber\\
 &=\diag(\Vb\Sigmab_p^{-1}\Vb^\top)^\top \diag(\Lambdab) \nonumber\\
&= \db_1\otimes\ldots\otimes\db_K \diag(\Lambdab)=\db_1\otimes\ldots\otimes\db_K \vect(\Dcal) \nonumber\\&
= \Dcal\times_1\db_1\ldots\times_K\db_K, \label{eq:second-relation}
\end{align}
where $\db_k = \diag(\Vb^{(k)}(\Sigmab^{(k)})^{-1}\Vb^{(k)^\top})^\top$ with $\Sigmab^{(k)}$ being a computed statistics in the E-step, $\diag(\Lambdab)$ denotes the diagonal elements of $\Lambdab$, and $\Dcal$ is a tensor of size $n_1\times\ldots\times n_K$, such that $\vect(\Dcal)=\diag(\Lambdab)$. Both time and space complexities of the last formula \eqref{eq:second-relation} is $O(\prod_{k=1}^K n_k)$.

We denote $\Vcal=[\Vb^{(1)},\ldots,\Vb^{(K)}]$ and $\Vcal^\top=[\Vb^{(1)\top},\ldots,\Vb^{(K)\top}]$.  For any tensor $\Acal$ of the same size as $\Dcal$, $\Lambdab \vect(\Acal)$ means multiplying the $j$-th element of $\vect(\Acal)$ by the $\Lambdab_{jj}$, which is the $j$-th element of $\vect{\Dcal}$. So we have $\Lambdab \vect(\Acal)=\vect(\Dcal\odot\Acal)$, where $\odot$ denotes the Hadamard product, \ie entry-wise product. In light of this relation, we can efficiently compute \eqref{eq:M-optimized} by
\begin{align} \label{eq:Upsilonb}
\hspace{-0.3cm}\Upsilonb\vect(\expec{q}{\Zcal}) = \vect\sbr{((\expec{q}{\Zcal}\times\Vcal^\top)\odot\Dcal)\times\Vcal}.
\end{align}
The right-hand side of Equation (\ref{eq:Upsilonb}) effectively reduce the time and space complexities of the left-hand side operations to $O(\sum_{k=1}^K n_k^3+(\sum_{k=1}^K n_k)\prod_{k=1}^K n_k)$ and $O(\sum_{k=1}^K n_k^2+\prod_{k=1}^K n_k)$, respectively.
%Relations similar to (\ref{eq:second-relation}) and (\ref{eq:Upsilonb}) also holds for the equations in the M-step and the prediction step.

We can further reduce the complexities by approximating the covariance matrices via truncated SVD. %If the first $t_k$ leading eigenvalues and the corresponding eigenvectors are preserved, the time complexity of our algorithm is $O(\sum_{k=1}^K n_k^2t_k+(\sum_{k=1}^K n_k)\prod_{k=1}^K t_k)$, and the space complexity is $O(\sum_{k=1}^K n_kt_k+\prod_{k=1}^K t_k)$.

%Previous literatures attack the high inference complexity problem by approximating the huge covariance matrix $\Sigma_p$ using the Nystr\"{o}m method or incomplete Cholesky factorization \citep{Zhang10multitask,Bonilla08multitask}. However, these methods can only be used in small-scale problems. For example, if we assume $K=4,~n_k=n$, the time and space complexities for exact GP inference is $O(n^{12})$ and $O(n^8)$, our algorithm reduce them to $O(n^5)$ time complexity and $O(n^4)$ space complexity. If we use $p=\sqrt{n}$ landmark points in Nystr\"{o}m and the first $\sqrt{n}$ leading eigenvalues and the corresponding eigenvectors in our algorithm, Nystr\"{o}m has time and space complexities of $O(n^8)$ and $O(n^6)$ respectively, but our algorithm only has $O(n^3)$ time complexity and $O(n^2)$ space complexity. Therefore, our algorithm can be applied to much larger problems.

\section{Related Works}
\label{sec:relatedworks}

 The \infiTucker model extends  Probabilistic PCA (PPCA) \citep{Tipping99PPCA} and Gaussian process latent variable models (GPLVMs) \citep{Lawrence06GPLVM}: while PPCA and GPLVM model interactions of one mode of a matrix and ignore the joint interactions of two modes, \infiTucker  does. Our model is also related to previous matrix-variate GPs \citet{Yu07SRM,Yu08GPlink}. The main difference lies in the fact they used linear covariance functions to reduce the computational complexities and dealt with matrix-variate data for online recommendation and link prediction.
%Recently \citet{Chu09ptucker} proposed a probabilistic Tucker-3 (pTucker) model \citep{Chu09ptucker}; our model reduces to the pTucker model when the degree of freedom $\nu\rightarrow\infty$.

The most closely related work is the probabilistic Tucker decomposition (pTucker) model \citep{Chu09ptucker}; actually the GP-based \infiTucker reduces to pTucker as a special case when using a linear covariance function. Our TP-based \infiTucker further differs from pTucker by marginalizing out a scaling hyperparameter of the covariance function. Another related work is probabilistic high order PCA \citep{Yu2011HOPCA}, which is essentially equivalent to a linear PPCA after transforming the tensor to a long vector.
 \citet{Hoff11multiway} proposed a hierarchical Bayesian extension to CANDECOMP/PARAFAC that captures the interactions of component matrices. Unlike these approaches,
  %both \citet{Chu09ptucker}'s and \citet{Hoff11multiway}'s approach,
  ours can handle non-Gaussian noise and uses nonlinear covariance functions to model complex interactions. In addition,  \citet{Chu09ptucker} did not exploits the Kronecker structure of the covariance matrices, so it is difficult for pTucker to scale to large datasets and high order tensors; and \citet{Hoff11multiway} used a Gibbs sampler for inference---requiring high computational cost and making their approach infeasible for tensors with moderate and large sizes.  By contrast, we provide a deterministic approximate inference method that exploits structures in Kronecker products in a variational Bayesian framework, making \infiTucker  much more efficient than competing methods.

To handle missing data, enhance model interpretability, and avoid overfitting, several extensions (e.g., using nonnegativity constraints) to tensor decomposition have been proposed, including nonnegative tensor decomposition (NTD) \citep{Shashua05NTF, KimChoi07NTD, Xiong10bptf, Porteous08MultiHDP} and Weighted tensor decomposition (WTD) \citep{Acar11missing}. Unlike ours, these models either solve the core tensors explicitly, or do not handle nonlinear multiway interactions.

Finally, note that the inference technique described in Section \ref{sec:inference} can be adopted for Gaussian process or $t$-process multi-task learning \citep{Bonilla08multitask,Zhang10multitask}. Let $M$ be the number of tasks and $N$ be the number of data points in each task. Our inference technique can be used to reduce their time complexity from $O(M^3N^3)$ to $O(M^3+N^3)$ and the space complexity from $O(M^2N^2)$ to $O(N^2+M^2)$.

\comment{

Our \infiTucker model is rooted in the probabilistic factor analysis models on matrices, \ie 2-mode tensors. Probabilistic PCA (PPCA) \citep{Tipping99PPCA} extracts latent component matrix for the rows and assumes independence across columns.
%Mathematically, the observation matrix $\Yb$ follows the matrix Gaussian distribution $\Ncal(\0,\{\Sigmab,\Ib\})$ in PPCA.
Gaussian process latent variable models (GPLVMs) \citep{Lawrence06GPLVM}, in contrary, model interactions of columns and assume independence of row.
%The observation matrix in GPLVM follows $\Ncal(\0,\{\Ib,\Sigmab\})$, where $\Sigmab$ is the covariance matrix of the component matrix $\Ub$.
%This modeling choice can be interpreted as a nonlinear mapping from latent feature space to output space.
Compared with \infiTucker, PPCA and GPLVM only model the interactions of one mode and ignore the potential correlation structures on the other mode.

Probabilistic methods that explicitly model correlations for each individual mode have been proposed in recent years. %Stochastic relational models
\citet{Yu07SRM,Yu08GPlink} describe stochastic relational processes (SRM) to approximate matrix-variate Gaussian processes with both row and column correlations. \citep{Yu08GPlink} proposed matrix Gaussian process models for link prediction and use linear covariance functions to reduce the computational complexities. Our model is mostly related to the probabilistic Tucker-3 (pTucker) model \citep{Chu09ptucker} and includes pTucker as a special case as $\nu\rightarrow\infty$. The critical differences are that our model uses tensor $t$ processes on general $K$-mode tensors and pTucker uses tensor Gaussian processes on 3-mode tensors. Our model allows non-Gaussian noise models and nonlinear covariance functions for component matrices, while pTucker cannot. Recently, \citet{Hoff11multiway} proposed a hierarchical Bayesian extension of CP that captures the interaction of component matrices via hierarchical modeling. However, the Gibbs sampler of \citep{Hoff11multiway} tends to converge slowly.

Other variants of tensor decomposition, including nonnegative tensor decomposition (NTD) and Weighted tensor decomposition (WTD) \citep{Acar11missing}, are also explored recently. NTD enforces nonnegativity constraints on the component matrices \citep{Shashua05NTF,KimChoi07NTD,Xiong10bptf,Porteous08MultiHDP}.
%\citep{Xiong10bptf} presented a Bayesian NTD model and applies it to temporal collaborative filtering. %\citep{Porteous08MultiHDP} proposed a probabilistic NTD model that extends latent Dirichlet allocation.
On the other hand, WTD tries to deal with missing value problems by penalizing the estimation
 %least square approximation
 errors of the observed data only. Different from \infiTucker, NTD and WTD models do not integrate out the core tensor, and there is no implicit feature mapping to model the nonlinear interactions.

The algebraic approach developed in section \ref{sec:inference} enables us to make efficient inference for nonlinear covariance functions. It is worth noting that the same inference strategy can be applied to Gaussian process or $t$-process multi-task learning \citep{Bonilla08multitask,Zhang10multitask,Yu07tmultitask}. Let $M$ be the number of tasks and $N$ be the number of data points in each task, our efficient inference strategy can be used to reduce their time complexity from $O(M^3N^3)$ to $O(M^3+N^3)$ and to reduce the space complexity from $O(M^2N^2)$ to $O(N^2+M^2)$.

} 
\section{Experiments}
\label{sec:experiment}

\begin{table}[!tb]
\centering
%\scriptsize
%\begin{small}
\small
\begin{tabular}{@{}l@{}|ccc}
\hline \textbf{Data} & amino & flow injection &bread  \\
\hline \hline
CP    &0.053$\pm$0.002   &0.051$\pm$0.005  &0.238$\pm$0.001	\\
TD    &0.054$\pm$0.002   &0.051$\pm$0.003  &0.248$\pm$0.001  \\
HOSVD &0.053$\pm$0.002   &0.052$\pm$0.004  &0.259$\pm$0.001 \\
NCP   &0.057$\pm$0.005   &0.110$\pm$0.023    &{0.233}$\pm$0.001 \\
PTD    &0.054$\pm$0.002  &0.048$\pm$0.002  &0.240$\pm$0.001\\
WCP   &0.049$\pm$0.004   &0.079$\pm$0.011       &0.246$\pm$0.003\\
\itgp & \textbf{0.047}$\pm$0.003  &{0.049}$\pm$0.002    & {0.232}$\pm$0.001\\
\ittp & \textbf{0.047}$\pm$0.003 &\textbf{0.046}$\pm$0.002 & \textbf{0.225}$\pm$0.001\\
 \hline
\end{tabular}
%\vspace{-0.2cm}
%\end{small}
\caption{\small The mean square errors (MSE) with standard errors.   The results suggested that our new approaches--\itgp and \ittp---achieved higher prediction accuracy than all the competing approaches. In particular, the improvements of \ittp over all the other methods on all datasets (except \itgp on the amino dataset) are statistically significant ($p<0.05$).}
\label{tab:res}
\end{table}

We use \itgp and \ittp to denote the two new infinite Tucker decomposition models based on tensor-variate Gaussian and $t$ processes, respectively.
To evaluate them, we conducted two sets of experiments, one on continuous tensor data and the other on binary tensor data. For both experiments, we compared \infiTucker with the following tensor decomposition methods: CANDECOMP/PARAFAC (CP), Tucker decomposition (TD),  Nonnegative CP (NCP), High Order SVD (HOSVD), Weighted CP (WCP) and Probabilistic Tucker Decomposition (PTD). We implemented PTD as described in the paper by  \citet{Chu09ptucker} and applied to a small continuous tensor data (\emph{bread} as described in the \ref{sec:exp1}). To handle larger and binary datasets, we used probit models and the efficient computation techniques described in Section \ref{sec:inf:comp} for PTD.
For the other methods, we used the implementation of the tensor data analysis toolbox\footnote{\small \url{http://csmr.ca.sandia.gov/~tgkolda/TensorToolbox/}} developed by T. G. Kolda.
%\begin{itemize}
%  \item CANDECOMP/PARAFAC (CP)
%  \item Tucker decomposition (TD)
%  \item None-Negative CP (NCP)
%  \item High Order Singular Value Decomposition (HOSVD)
%  \item Weighted CP (WCP)
%\end{itemize}

%In the following, we introduce the datasets, experimental setting, and experimental results for two batches of experiments, respectively.
\subsection{Experiment on continuous tensor data}
%\textbf{Experiment on Gaussian Noise Model}
\subsubsection{Experimental setting}\label{sec:exp1}
%\textbf{Experimental setting:} %{To examine the performance of the Tensor TP algorithm with Gaussian noise,
We used three continuous chemometrics datasets\footnote{\small Available from \url{http://www.models.kvl.dk/datasets}}, \emph{amino}, \emph{bread}, and \emph{flow injection}.
The  \emph{amino} dataset consists of five simple laboratory-made samples. Each sample contains different amounts of tyrosine, tryptophan and phenylalanine dissolved in phosphate buffered water. The samples were measured by fluorescence (excitation 250-300 nm, emission 250-450 $nm$, 1 nm intervals) on a PE $LS50B$ spectrofluorometer with excitation slit-width of 2.5 $nm$, an emission slit-width of 10 $nm$ and a scan-speed of 1500 $nm$/$s$. Thus the dimension of the tensor is  5 $\times$ 51 $\times$ 201. The \emph{bread} data describes five different breads which were baked in replicates, giving a total of ten samples. Eight different judges assessed the breads with respect to eleven different attributes in a fixed vocabulary profiling analysis. Hence the dimension of the tensor is 10 $\times$ 11 $\times$ 8. The \emph{flow injection} data describes a flow injection analysis (FIA) system where a pH-gradient is imposed. In this setup, a carrier stream containing a Britton-Robinson buffer of $pH$ 4.5 is continuously injected into the system with a flow of 0.375 mL/min. The 77 $\mu L$ of sample and 770 $\mu L$ of reagent (Britton-Robinson buffer $pH$ 11.4) are injected simultaneously into the system by a six-port valve and the absorbance is detected by a diode-array detector (HP 8452A) from 250 to 450 $nm$ in two nanometer intervals. The absorption spectrum is determined every second 89 times during one injection. Thus this dataset is a 12 (samples) $\times$ 100 (wavelengths) $\times$ 89 (times) array.

%Their dimensions are $5\times 201\times 61$, $10\times 11\times 8$, and $12\times 100\times 89$, respectively.
 %It should be noted
%that the number of edges in a tensor is in a polynomial order of
%the number of nodes, and the prediction is made on each edge.
%The large number of edges makes the prediction task computationally challenging.
All the above tensor data were normalized such that each element of the tensor has zero mean and unit variance (based on the vectorized representations).
% their Kronecker products over all modes have zero mean and unite variance.
For each tensor, we randomly split it via 5-fold cross validation: each time four folds are used for training and one fold for testing. This procedure was repeated 10 times, each time with a different partition for the 5-fold cross validation. In \ittp, the degree of freedom $\nu$ in the tensor-variate $t$ process is fixed to $10$. We chose the Gaussian/exponential covariance functions $\Sigma^{(k)}(\uu_i,\uu_j)=e^{-\gamma\|\uu_i-\uu_j\|^t}$, where $t=1,2$ and $\gamma$ is selected from $[0.01:0.05:1]$ by 5-fold cross validation. The regularization parameter $\lambda$ for \itgp and \ittp is chosen from $\{1, 10, 100\}$.

\subsubsection{Results}
%\textbf{Results:}
We  compared the the prediction accuracies of all the approaches  on hold-out elements of the tensor data.  For each comparison, we used the same number of latent factors,  denoted as $r$,  for all the approaches. We varied $r$ from 3 to 5 and computed the averaged mean square errors (MSEs) and the standard errors of the MSEs. Based on cross-validation, we set  $r=3$. The MSEs on the three datasets are summarized in Table \ref{tab:res}.
%, where the lowest MSE value and those not significantly worse than it (achieved by t-test with $95\%$ confidence level) are highlighted.
Based on the prediction accuracies, PTD and WCP tie on the third best, while HOSVD is the worst ( perhaps due to the strong nonnegativity constraint on the latent factors).
Clearly, \itgp achieved higher prediction accuracies than all the previous approaches on all the datasets, and \ittp further outperformed \itgp for most cases. %This may be due to the so-called ``kernel trick''  that the non-linear covariance functions provide more flexibility in modeling the interaction among all modes.

\begin{figure}[!ht]
\centering
\subfigure[\emph{Enron}]{\includegraphics[width=.5\textwidth]{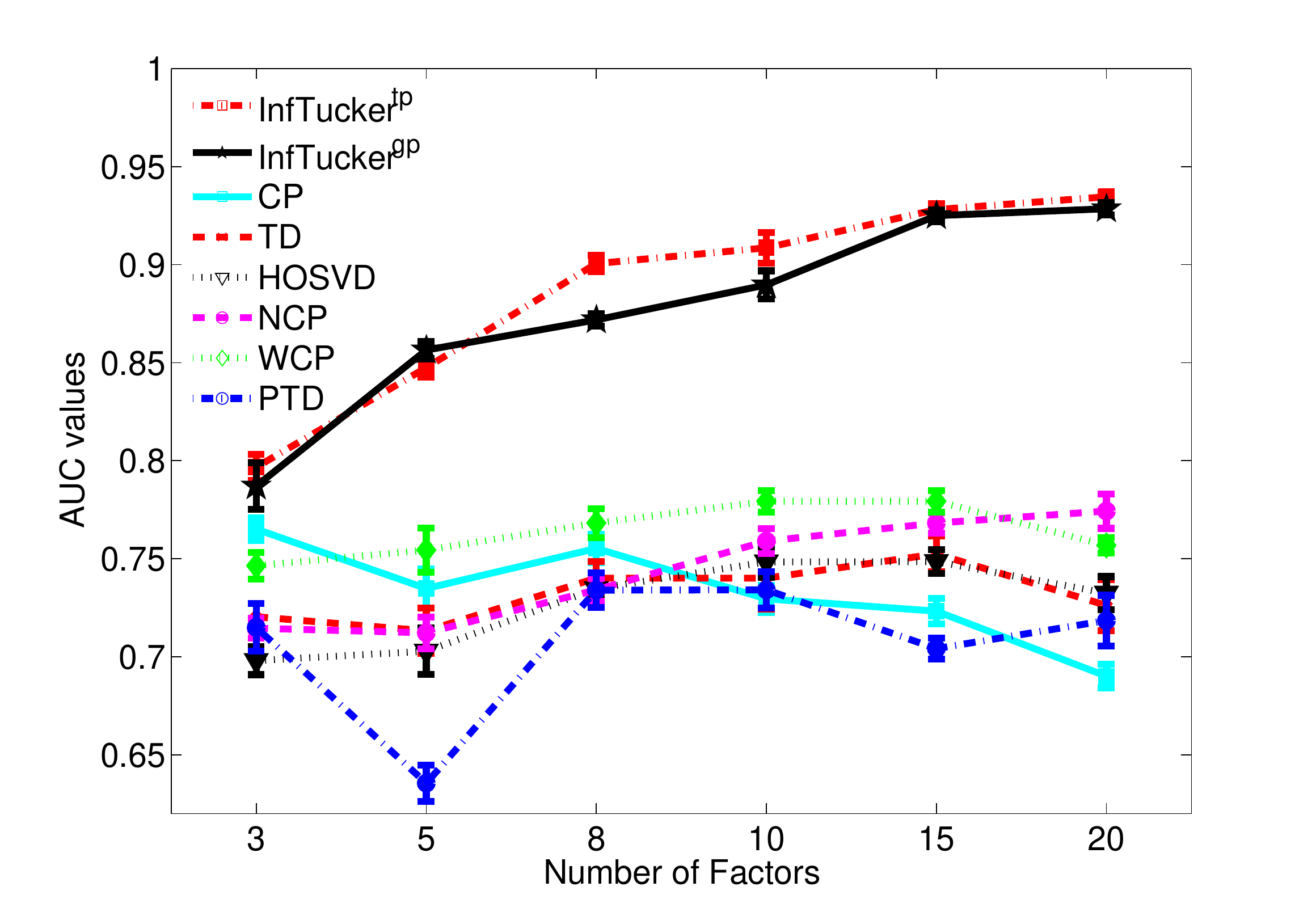}}
\subfigure[\emph{digg1}]{\includegraphics[width=.5\textwidth]{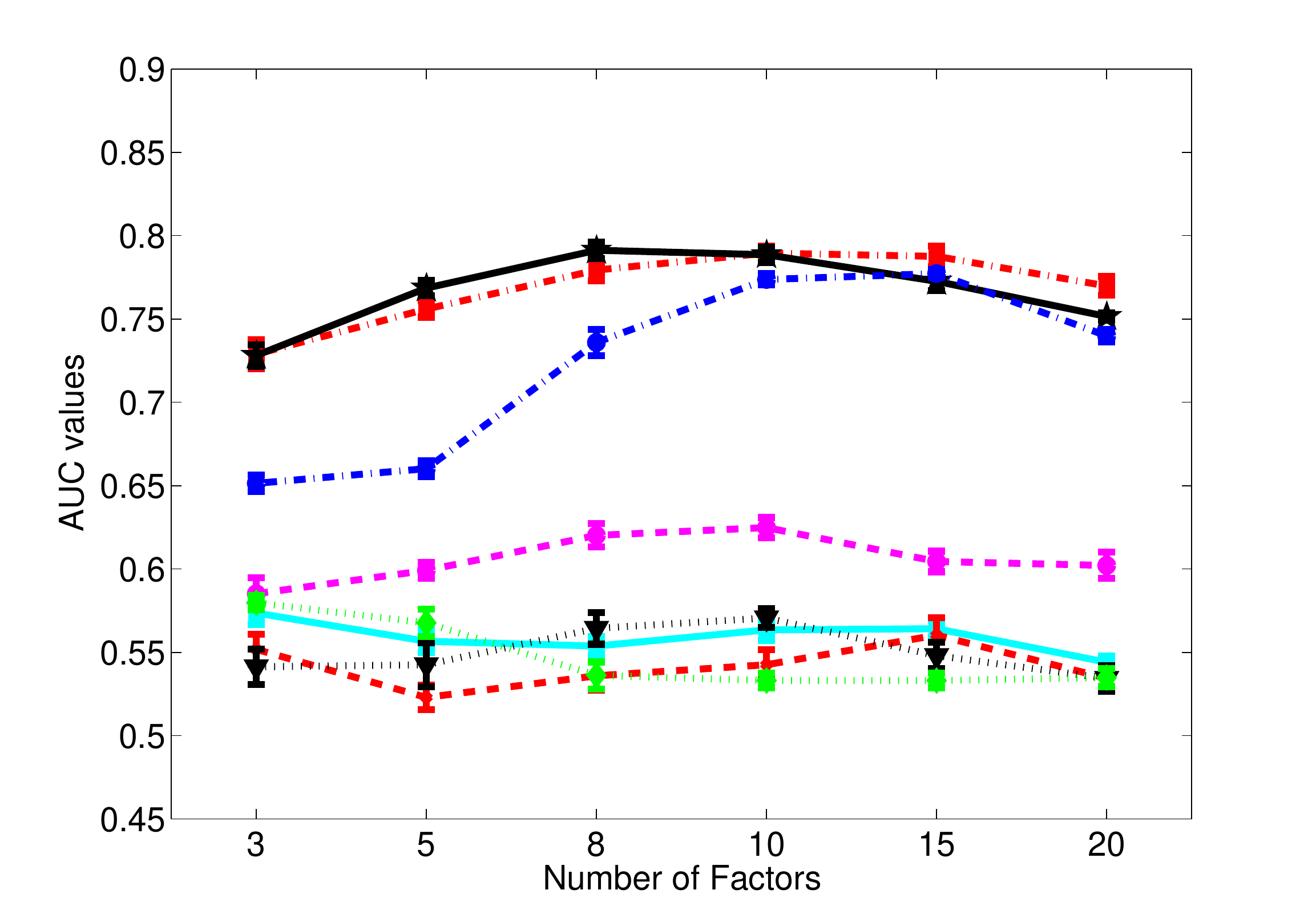}}
\subfigure[\emph{digg2}]{\includegraphics[width=.5\textwidth]{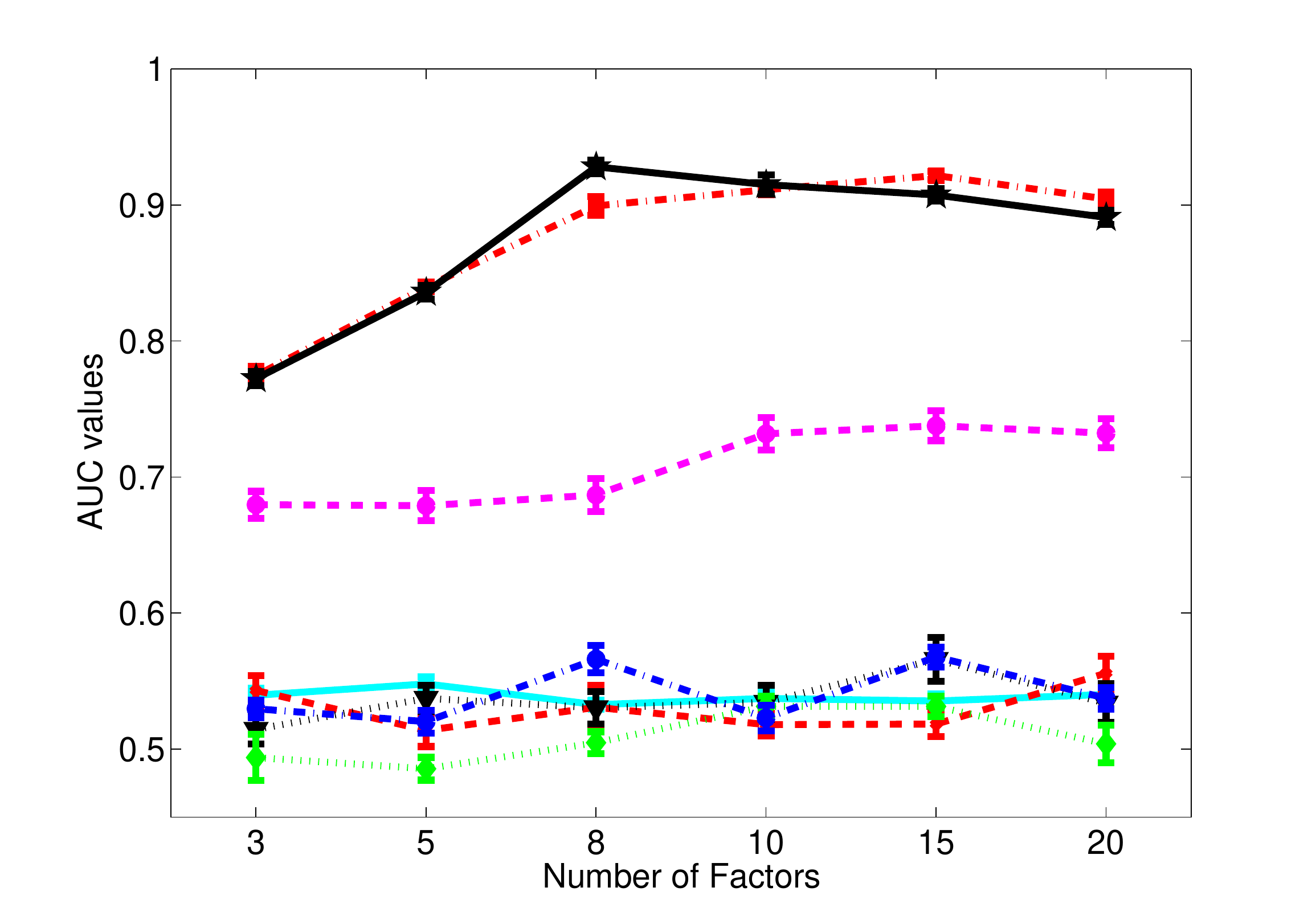}}
 \caption{The area under curve (AUC) values of six algorithms on three multi-way networks. The dimensions of the latent factors are $r=3,5,8,10,15,20$ respectively. The proposed \InfTucker models performed significantly better than the other methods.
 %\infiTucker outperforms the other algorithms significantly on all datasets.
 %Individual figures in original sizes are provided in the appendix.
 }
\label{fig:auc}
\end{figure}

%\begin{figure*}[ht]
%\vspace{-0.1cm}
%\centering
%\hspace{-0.4cm}\subfigure[\emph{Enron}]{\includegraphics[width=0.33\textwidth]{./fig/enron.pdf}}
%\hspace{-0.5cm}\subfigure[\emph{digg1}]{\includegraphics[width=0.33\textwidth]{./fig/digg1.pdf}}
%\hspace{-0.5cm}\subfigure[\emph{digg2}]{\includegraphics[width=0.33\textwidth]{./fig/digg2.pdf}}
% \caption{\small The AUC values of six algorithms on three multi-way networks. The dimensions of the latent factors are $r=3,5,8,10,15,20$ respectively.
% %\infiTucker outperforms the other algorithms significantly on all datasets.
% %Individual figures in original sizes are provided in the appendix.
% }
%\vspace{-0.1cm}
%\label{fig:auc}
%\end{figure*}

\subsection{Experiment on binary tensor data}

%\textbf{Experiment on Probit Noise Model}
\subsubsection{Experimental setting}
%{\bf Experimental setting:} %In order to evaluate the \infiTucker model with probit noise,
We extracted three binary social network datasets,  \emph{Enron}, \emph{Digg1}, and \emph{Digg2},  for our experimental evaluation. \emph{Enron} is a relational dataset describing the three-way relationship: sender-receiver-email. This dataset, extracted from the Enron email dataset\footnote{\small Available at \url{http://www.cs.cmu.edu/~enron/}}, has the dimensionality of $203\times 203\times 200$ with 0.01\% non-zero elements. The \emph{Digg1} and \emph{Digg2} datasets were all extracted from a social news website \url{digg.com}\footnote{\small Available at \url{http://www.public.asu.edu/~ylin56/kdd09sup.html}}. \emph{Digg1} describes a three-way interaction: news-keyword-topic, and \emph{Digg2} describes a four-way interaction: user-news-keyword-topic. \emph{Digg1} has the dimensionality of $581\times124\times48$ with 0.024\% non-zero elements, and \emph{Digg2} has the dimensionality of $22\times 109\times 330\times 30$ with only 0.002\% non-zero elements. Apparently these datasets are very sparse.
%{\bf Experimental setting:} We adopt a similar settings to the Gaussian noise case.

\subsubsection{Results}
%{\bf Results:}
We chose $r$ from the range \{3,5,8,10,15,20\} based on cross-validation. Since the data are binary,  we evaluated all these approaches by area-under-curve (AUC) values averaged over 50 runs. The larger the averaged AUC value an approach achieves, the better it is. %In \infiTucker, we use $\expec{q}{\Mcal}$ as predictions to calculate %the AUC values.
We reported the averaged AUC values for all algorithms in Figure \ref{fig:auc}. Again, the proposed \itgp and \ittp approaches significantly outperform all the others. Note that the  nonprobabilistic approaches---such as CP and TD---- suffer severely from the least square minimization; given the sparse and binary training data, the least-square-minimization leads to too many predictions with zero values, a result of both overfitting and mis-model fitting. This experimental comparison fully demonstrates the advantages of \infiTucker (stemming from the  right noise models and the nonparametric Bayesian treatment).
\section{Conclusion}

To conduct multiway data analysis, we have proposed a new nonparametric Bayesian tensor decomposition framework, \infiTucker, where the observed tensor is modeled as a sample from a stochastic processes on tensors. In particular, we have employed tensor-variate Gaussian and $t$ processes.
This new framework can model nonlinear interactions between multi-aspects of the tensor data, handle missing data and noise, and quantify prediction confidence (based on predictive posterior distributions). We have also presented an efficient variational method to estimate \infiTucker from data. Experimental results on chemometrics and social network datasets demonstrated that the superior predictive performance of \infiTucker over the alternative tensor decomposition approaches.

%\noindent{\bf{Acknowledgment}}
%
%We thank the generous support from A, B, and C.

% \pagebreak 
\newpage
\begin{small}
\bibliography{literature}
\bibliographystyle{plainnat}
\end{small}

\newpage
\appendix

\section{Proof of Theorem \ref{thm:tensor-process-convergence}} \label{subsec:proof}

\begin{proofsketch}
If $\Wcal^{(r)} \sim \TN(\0,\{\Ib_r\}_{k=1}^K)$, then $\vect(\Wcal^{(r)}) \sim \Ncal(\0,\Ib_{rK})$. Let $\Ub^{(k)},~k=1,\ldots,K$ be $K$ location sets (matrices) as used in \eqref{eq:infinite-tucker-limit}, we have
\begin{align}
\label{eq:feature-space-identity}
\vect\rbr{\Wcal^{(r)}\times\phi^{(r)}(\Ucal)} = \phi^{(r)}(\Ub^{(1)})\otimes\ldots\otimes\phi^{(r)}(\Ub^{(K)})\vect(\Wcal^{(r)})
\end{align}
Thus, $\vect\rbr{\Wcal^{(r)}\times\phi^{(r)}(\Ucal)} \sim \Ncal(\nu,\0,\Sigmab^{(1)}_r\otimes\ldots\otimes\Sigmab^{(K)}_r)$, where $\Sigmab^{(k)}_r(i,j) = \Sigma^{(k)}_r(\uu^{(k)}_i,\uu^{(k)}_j)$ are the covariance matrix. Inverting \eqref{eq:feature-space-identity} gives $\Wcal^{(r)}\times\phi^{(r)}(\Ucal) \sim \TN(\0,\{\Sigmab^{(k)}_r\}_{k=1}^K)$, which proves $t^{(r)}$ follows the tensor Gaussian process.

From the definition of inner product in $\ell_2$, we have the following identity on the convergence of covariance function. $$\Sigma^{(k)}(\uu^{(k)}_i,\uu^{(k)}_j)=\lim_{r\rightarrow\infty}\Sigma^{(k)}_r(\uu^{(k)}_i,\uu^{(k)}_j),~\forall\uu^{(k)}_i,\uu^{(k)}_j\in U_k$$
Convergence in distribution follows from this convergence result.

%It is well known that $t$ distribution can be written in the form
%\begin{align}
%\label{eq:t-distribution-fac}
%\Tcal(\xb|\nu,\mub,\Sigmab)=\int\Ncal(\xb|\mub,\eta^{-1}\Sigmab)\textrm{Gam}(\eta|\nu/2,\nu/2) d\eta
%\end{align}
%where $\rm{Gam}(\eta|\gamma,\rho)$ denotes Gamma distribution with shape parameter $\gamma$ and scale parameter $\rho$. Given the Gamma random varaiable $\eta$, the claim for tensor $t$ process reduces to the case of tensor Gaussian process. Then it can be proved by integrating out $\eta$.
\end{proofsketch}

\section{Gradient of $f(\Ucal)$} \label{subsec:grad}

\begin{align}
\frac{\partial{f}}{\partial{u^{(k)}_{ij}}} = \;&\frac{n}{n_k} \tr\rbr{(\Sigmab^{(k)})^{-1} \frac{\partial\Sigmab^{(k)}}{\partial{u^{(k)}_{ij}}}} + \tau \mub^\top\Deltab^{(k)}\mub + \tau \tr\rbr{\Deltab^{(k)}\Upsilonb} \\
\Deltab^{(k)} = \;&(\Sigmab^{(1)})^{-1}\otimes\ldots\otimes(\Sigmab^{(k-1)})^{-1}\otimes (\Sigmab^{(k)})^{-1} \frac{\partial\Sigmab^{(k)}}{\partial{u^{(k)}_{ij}}} (\Sigmab^{(k)})^{-1} \nonumber\\ &\otimes(\Sigmab^{(k+1)})^{-1}\otimes\ldots\otimes(\Sigmab^{(K)})^{-1} \nonumber
\end{align}

\end{document}